\documentclass[10pt,twocolumn,letterpaper]{article}

 \usepackage[pagenumbers]{wacv} 
\usepackage[accsupp]{axessibility}
\usepackage{epsfig}
\usepackage{graphicx}
\usepackage{amsmath}
\usepackage{amssymb}
\usepackage{enumitem}
\usepackage{tikz}
\usetikzlibrary{shapes.geometric, arrows.meta, fit, backgrounds, calc}
\definecolor{wacvblue}{rgb}{0.21,0.49,0.74}
\usepackage[breaklinks,colorlinks,allcolors=wacvblue]{hyperref}

\def\wacvPaperID{*****} 
\def\confName{WACV}
\def\confYear{2027}

\title{Validation-Guided Prompt Selection for Differential Morphing Attack Detection}

\author{
Raghavendra Ramachandra \qquad Hailin Li\\
SAFE Centre, Norwegian University of Science and Technology (NTNU)\\
Gj{\o}vik, Norway\\
{\tt\small
raghavendra.ramachandra@ntnu.no \qquad
hailin.li@ntnu.no
}
}

\begin{document}
\maketitle
\begin{abstract}
Differential morphing attack detection (D-MAD) compares a trusted live capture with a document image that may contain a facial morph. We investigate whether labelled validation data can improve a frozen-parameter vision-language evaluator through prompt selection rather than model fine-tuning. Version~1 generates 50 candidate prompts and selects the top-ranked instruction using a security-oriented loss that prioritises missed morph attacks (MACER) while accounting for bona fide classification errors (BSCER), response failures, parse failures, and decision inconsistency. Version~2 recursively revises prompts using aggregate validation feedback. On a subject-disjoint test set of 160 pairs spanning LMA, MIPGAN-2, and PIPE, the selected Version~1 prompt achieves a (19.17\%) morph attack classification error rate (MACER), with (0.00\%) observed bona fide sample classification error rate (BSCER), compared with (70.00\%) MACER for a controlled manual prompt. The paired (50.83) percentage-point difference has a (95\%) subject-bootstrap interval of ([43.33,58.33]). Recursive revision does not surpass the best Generation~0 validation loss; the reported revised descendant obtains (31.67\%) test MACER. All 23 attacks missed by Version~1 are landmark-based (LMA) morphs. Overall, the results show that prompt formulation can substantially influence frozen-parameter VLM-based D-MAD performance, while also highlighting the challenge of recursive prompt refinement under a compact 10-subject validation setting.

\end{abstract}

\section{Introduction}
\label{sec:intro}

Face verification is central to Automated Border Control (ABC), where a live
capture is compared with the face image stored in an electronic
machine-readable travel document~\cite{Manasa2025TimFusNet}. This process
assumes that the document image represents one genuine identity. Face morphing
attacks violate this assumption by combining information from two contributors,
allowing both to resemble the document photograph during enrolment or later
verification~\cite{DiDomenico2025ACIdA,Shekhawat2025}. Such attacks can be
produced through landmark-based blending~\cite{Landmark-face-morph}, GAN-based
synthesis~\cite{zhang-MIPGAN-TBIOM-2021}, and diffusion-based
generation~\cite{PIPE}.

Single-image MAD examines a document image in isolation, whereas Differential
MAD (D-MAD) compares it with a trusted live capture. The differential setting
provides identity evidence that is unavailable in single-image inspection and
therefore fits the ABC workflow. Most D-MAD methods nevertheless require
labelled morph pairs to train a classifier or adapt a pretrained
representation~\cite{Banas2026IDSwapMAD,GonzalezSoler2026DifFoundMAD}. Their
development is constrained by limited biometric data and by the difficulty of
covering morph generation and acquisition conditions that may arise after
deployment.

A frozen-parameter Vision-Language Model (VLM) offers a complementary approach
because it can compare two facial images without model fine-tuning. Existing
studies demonstrate zero-shot MAD and related presentation attack detection
with manually defined prompts~\cite{Shekhawat2025,Zhang2026ChatGPT,MADCLIP,%
Komaty2025ChatGPTPAD}. Prompt choice therefore remains an uncontrolled
experimental factor. Within VLM-based MAD, no subject-disjoint validation rule
has been studied for selecting among alternatives. Our procedure uses
validation labels and is supervised prompt selection, although no model
parameters are updated.

We treat prompt design as a controlled D-MAD variable. Version~1 generates 50
candidates, ranks them on a validation split with disjoint subjects, and fixes
the top-ranked prompt before testing. Version~2 starts from 30 Generation~0
prompts and revises selected candidates using aggregate validation feedback.
Both variants minimise the security loss \(\mathcal{L}_{\mathrm{sec}}\), which
prioritises missed attack reduction while accounting for bona fide errors and
response reliability. The evaluator configuration is held constant within each
comparison, and no test image, label, or metric contributes to prompt
generation or selection.

The main contributions of this work are:
\begin{enumerate}[leftmargin=*,nosep,topsep=2pt,label=\arabic*.]
    \item We instantiate prompt design as a validation-guided selection problem
    and compare a manual prompt, non-recursive selection, and recursive
    optimisation under the same frozen-parameter evaluator configuration and
    test protocol.

    \item We define a failure-aware evaluation protocol with MACER-oriented
    prompt ranking, OR decision fusion, maximum morph-risk-score fusion, and
    explicit handling of response and parse failures.

    \item We evaluate subject-disjoint bona fide and morph pairs covering different morphing generation techniques such as LMA \cite{Ferrara-TextureBlendingAndShapeWarpingInFaceMorphing-IEEE-BIOSIG-2019},
    MIPGAN-2 \cite{zhang-MIPGAN-TBIOM-2021}, and PIPE \cite{PIPE}, with analyses of morph-specific errors, response budget,
    subject-level uncertainty, and recursive optimisation behaviour.
\end{enumerate}

\section{Related Work}
\label{sec:rel}

\noindent\textbf{Conventional D-MAD.}
Early differential approaches compare facial representations extracted
from the document image and live capture. The resulting feature difference
or similarity is evaluated directly or passed to a lightweight classifier
such as an SVM~\cite{Scherhag-FaceMorphingAttacks-TIFS-2020,
10.1371/journal.pone.0304610,Manasa2025TimFusNet}. These methods benefit
from identity information in the trusted reference image, but their
performance depends on the representation model, the training attacks,
and the availability of labelled morph pairs. De-morphing methods instead
estimate and remove the live capture contribution before testing whether
the residual is compatible with another identity
~\cite{Ferrara-Demorphing-TIFS-2018,Hoang2025Demorph}. Their behaviour can
depend on assumptions about the blending process and the quality of the
input images.

\medskip
\noindent\textbf{Foundation models for MAD.}
Pretrained foundation models reduce the need to learn representations from
limited biometric data. MADation~\cite{caldeira2025madationfacemorphingattack}
and FoundPAD~\cite{ozgur2025foundpad} adapt CLIP with LoRA for MAD/PAD, while
DifFoundMAD~\cite{GonzalezSoler2026DifFoundMAD} uses differential DINO features
with a lightweight trainable component. These methods require labelled
adaptation data. Training-free studies instead use image-text or multimodal
language models with predefined prompts for MAD and related biometric attack
detection~\cite{MADCLIP,Shekhawat2025,Zhang2026ChatGPT,Komaty2025ChatGPTPAD}.


\noindent\textbf{Prompt design in biometric security.}
Recent work explores prompting multimodal models for biometric tasks. Zhang
et al.~\cite{Zhang2026ChatGPT} use designed prompts for zero-shot MAD;
Shekhawat et al.~\cite{Shekhawat2025} extend this to D-MAD with
chain-of-thought instructions; Patwardhan et al.~\cite{MADCLIP} compare
manually written prompts; and Narayan et al.~\cite{Narayan2026FaceXBench}
evaluate zero-shot, task-description, and chain-of-thought prompting across
face tasks. These studies show prompt sensitivity but do not select an
instruction using subject-disjoint labelled validation data. General prompt
optimisation generates, selects, or iteratively refines instructions
~\cite{Zhou2023APE,Yang2024OPRO,Qu2025ProAPO}. We instead treat prompt choice
as a controlled D-MAD variable: candidates are ranked on subject-disjoint
validation data and frozen before test evaluation. Thus, our contribution is
supervised black-box prompt selection for D-MAD, not a new general
prompt-optimisation algorithm.

\section{Proposed Method}
\label{sec:method}
This section formulates the D-MAD task and introduces the evaluator schema and prompt-selection framework used throughout the study.

\subsection{D-MAD Formulation and Evaluator Schema}
\label{sec:setting}

A D-MAD sample is represented by an image pair
\(\mathcal{X}_{i}=(A_{i},B_{i})\), where \(A_{i}\) is a trusted bona fide
reference or live capture image and \(B_{i}\) is a document image presented
for verification. The ground truth label is \(y_{i}\in\{0,1\}\), with
\(y_{i}=0\) denoting a bona fide pair and \(y_{i}=1\) denoting a morphing
attack. The objective is to predict whether \(B_{i}\) is a morphing attack
relative to \(A_{i}\). The morph generation method is not supplied to the
evaluator or to any agent in the framework.

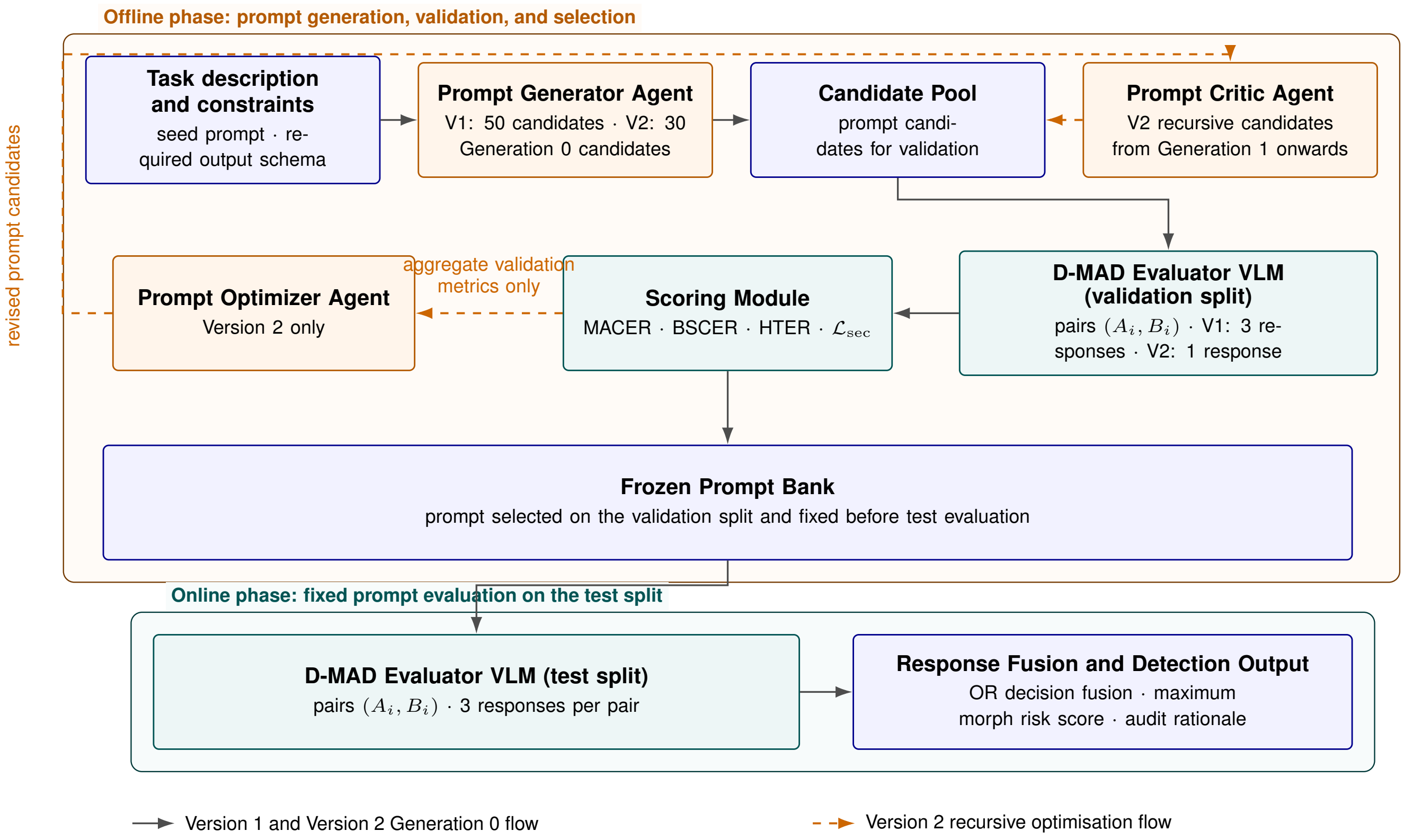
\begin{figure*}[t]
  \centering
  \begingroup
  \tikzset{
    dmadData/.style={
      draw=blue!55!black, fill=blue!5, rounded corners=2pt,
      align=center, font=\footnotesize\sffamily,
      inner sep=5pt, line width=0.55pt, minimum height=1.48cm},
    dmadAgent/.style={
      draw=orange!70!black, fill=orange!8, rounded corners=2pt,
      align=center, font=\footnotesize\sffamily,
      inner sep=5pt, line width=0.55pt, minimum height=1.48cm},
    dmadEval/.style={
      draw=teal!65!black, fill=teal!7, rounded corners=2pt,
      align=center, font=\footnotesize\sffamily,
      inner sep=5pt, line width=0.55pt, minimum height=1.48cm},
    dmadOffline/.style={
      draw=orange!45!black, fill=orange!3, rounded corners=4pt,
      inner sep=8pt, line width=0.45pt},
    dmadOnline/.style={
      draw=teal!45!black, fill=teal!3, rounded corners=4pt,
      inner sep=8pt, line width=0.45pt},
    dmadArrow/.style={
      -{Latex[length=2.2mm,width=1.5mm]},
      line width=0.65pt, draw=black!70},
    dmadVTwoArrow/.style={
      -{Latex[length=2.2mm,width=1.5mm]},
      line width=0.75pt, draw=orange!80!black, dashed},
    dmadTag/.style={
      font=\scriptsize\sffamily, color=orange!80!black,
      inner sep=1.5pt, anchor=south},
    dmadPhase/.style={
      font=\scriptsize\sffamily\bfseries, inner sep=2pt,
      anchor=south west}
  }

  \resizebox{\textwidth}{!}{%
  \begin{tikzpicture}

  \node[dmadData, text width=3.45cm] (task) at (2.95,9.15)
    {\textbf{Task description and constraints}\\[1pt]
     \scriptsize seed prompt $\cdot$ required output schema};

  \node[dmadAgent, text width=3.45cm] (generator) at (7.25,9.15)
    {\textbf{Prompt Generator Agent}\\[1pt]
     \scriptsize V1: 50 candidates $\cdot$ V2: 30 Generation~0 candidates};

  \node[dmadData, text width=3.45cm] (pool) at (11.55,9.15)
    {\textbf{Candidate Pool}\\[1pt]
     \scriptsize prompt candidates for validation};

  \node[dmadAgent, text width=3.45cm] (critic) at (15.85,9.15)
    {\textbf{Prompt Critic Agent}\\[1pt]
     \scriptsize V2 recursive candidates from Generation~1 onwards};

  \node[dmadEval, text width=5.05cm] (vlmval) at (15.05,6.65)
    {\textbf{D-MAD Evaluator VLM}\\[-1pt]
     \textbf{(validation split)}\\[1pt]
     \scriptsize pairs $(A_i,B_i)$ $\cdot$ V1: 3 responses
     $\cdot$ V2: 1 response};

  \node[dmadEval, text width=3.90cm] (scoring) at (9.35,6.65)
    {\textbf{Scoring Module}\\[1pt]
     \scriptsize MACER $\cdot$ BSCER $\cdot$ HTER
     $\cdot$ $\mathcal{L}_{\mathrm{sec}}$};

  \node[dmadAgent, text width=3.55cm] (optimizer) at (3.35,6.65)
    {\textbf{Prompt Optimizer Agent}\\[1pt]
     \scriptsize Version~2 only};

  \node[dmadData, text width=15.80cm] (frozen) at (9.35,4.20)
    {\textbf{Frozen Prompt Bank}\\[1pt]
     \scriptsize prompt selected on the validation split and fixed before test evaluation};

  \node[dmadEval, text width=8.00cm] (vlmtest) at (6.10,1.75)
    {\textbf{D-MAD Evaluator VLM (test split)}\\[1pt]
     \scriptsize pairs $(A_i,B_i)$ $\cdot$ 3 responses per pair};

  \node[dmadData, text width=6.10cm] (output) at (14.20,1.75)
    {\textbf{Response Fusion and Detection Output}\\[1pt]
     \scriptsize OR decision fusion $\cdot$ maximum morph risk score
     $\cdot$ audit rationale};

  \begin{scope}[on background layer]
    \node[dmadOffline,
      fit=(task)(generator)(pool)(critic)(vlmval)(scoring)(optimizer)(frozen)]
      (offline) {};
    \node[dmadOnline, fit=(vlmtest)(output)] (online) {};
  \end{scope}

  \node[dmadPhase, color=orange!70!black, fill=orange!3]
    at ($(offline.north west)+(0.45,0)$)
    {Offline phase: prompt generation, validation, and selection};
  \node[dmadPhase, color=teal!65!black, fill=teal!3]
    at ($(online.north west)+(0.45,0)$)
    {Online phase: fixed prompt evaluation on the test split};

  \draw[dmadArrow] (task.east) -- (generator.west);
  \draw[dmadArrow] (generator.east) -- (pool.west);
  \draw[dmadArrow] (pool.south) -- ++(0,-0.28) -| (vlmval.north);
  \draw[dmadArrow] (vlmval.west) -- (scoring.east);
  \draw[dmadArrow] (scoring.south) -- (frozen.north);
  \draw[dmadArrow] (frozen.south) -- ++(0,-0.32) -| (vlmtest.north);
  \draw[dmadArrow] (vlmtest.east) -- (output.west);

  \draw[dmadVTwoArrow] (scoring.west) -- (optimizer.east);
  \node[dmadTag, text width=2.35cm, align=center]
    at ($(scoring.west)!0.5!(optimizer.east)+(0,0.16)$)
    {aggregate validation\\metrics only};

  \draw[dmadVTwoArrow] (optimizer.west) -- (0.75,6.65) -- (0.75,10.00)
    -- (15.85,10.00) -- (critic.north);
  \node[dmadTag, rotate=90, anchor=south] at (0.30,7.65)
    {revised prompt candidates};

  \draw[dmadVTwoArrow] (critic.west) -- (pool.east);

  \draw[dmadArrow] (1.65,0.05) -- ++(0.55,0)
    node[right, font=\scriptsize\sffamily]
    {Version~1 and Version~2 Generation~0 flow};
  \draw[dmadVTwoArrow] (10.45,0.05) -- ++(0.55,0)
    node[right, font=\scriptsize\sffamily]
    {Version~2 recursive optimisation flow};

  \end{tikzpicture}%
  }
  \endgroup

  \caption{Agentic prompt generation framework for D-MAD. Version~1 and
  Version~2 Generation~0 follow the solid path from prompt generation to
  validation. Recursive Version~2 candidates follow the dashed path from the
  Scoring Module through the Prompt Optimizer and Prompt Critic before returning
  to the Candidate Pool. The selected prompt is fixed before test evaluation.
  Test responses are subsequently combined by deterministic decision and score
  fusion.}
  \label{fig:dmad_pipeline}
\end{figure*}

GPT-4o~\cite{OpenAI2024gpt4o} serves as the D-MAD Evaluator VLM shown in
Fig.~\ref{fig:dmad_pipeline}. Its multimodal input support and structured
outputs allow prompt formulations to be compared under a common evaluator
configuration. ``Frozen'' denotes that the authors perform no fine-tuning or
parameter updates; it does not imply control over provider-side model updates.
For a prompt \(p\), image pair \(\mathcal{X}_{i}\), and response index \(r\),
the evaluator produces
\begin{equation}
    \mathcal{Z}_{i,r}(p)
    =
    \mathcal{E}_{\theta}
    \bigl(A_{i},B_{i},p\bigr),
    \label{eq:vlm_evaluator}
\end{equation}
where \(\mathcal{E}_{\theta}\) denotes an evaluator that receives no
author-controlled parameter updates. All evaluator
queries use a temperature of zero. Prompt generation and recursive prompt
revision use separate settings described in Sections~\ref{sec:agents}
and~\ref{sec:v2}.

The evaluator returns a structured JSON response containing a same-identity
assessment, an identity confidence score, a binary morph detection decision, a
morph risk score on a \(0\)--\(100\) scale, up to three items of observable
evidence, an image-quality caveat, and a decision reliability level. The morph
detection field defines the response decision
\(\hat{y}_{i,r}\in\{0,1\}\), and the morph risk value is rescaled to
\(s_{i,r}\in[0,1]\). We refer to this value as a morph risk score because it
is not a calibrated probability; it is used only for ranking, threshold
analysis, and EER computation. The evidence is retained as a model-generated
rationale, not as a verified causal explanation.

The experiments consider landmark-based morphing
(LMA)~\cite{Ferrara-TextureBlendingAndShapeWarpingInFaceMorphing-IEEE-BIOSIG-2019,%
Landmark-face-morph}, MIPGAN-2~\cite{zhang-MIPGAN-TBIOM-2021},
and the diffusion-based PIPE method~\cite{PIPE}. These technique labels are
used only after inference to report performance by morphing method. They are
not included in the task prompt, Candidate Pool, Frozen Prompt Bank, or
Detection Output.

For a set of morphing attack pairs \(\mathcal{M}\) and bona fide pairs
\(\mathcal{B}\), the primary decision error rates are the Morphing Attack
Classification Error Rate (MACER) and the Bona Fide Sample Classification
Error Rate (BSCER). We use BSCER rather than the presentation-oriented BPCER,
following morphing attack evaluations such as FVC-onGoing and NIST FATE
MORPH~\cite{FVC-Ongoing-DMAD-Evaluation,Nist-Frvt-Morph}, because a D-MAD
decision is taken on a stored document sample rather than on a live
presentation at a capture device, which also keeps the terminology consistent
with MACER. They are defined as
\begin{equation}
    \mathrm{MACER}
    =
    \frac{1}{|\mathcal{M}|}
    \sum_{i\in\mathcal{M}}
    \mathbb{I}
    \left[
        \hat{y}^{\,\mathrm{fused}}_{i}\in\{0,\bot\}
    \right],
    \label{eq:macer}
\end{equation}
and
\begin{equation}
    \mathrm{BSCER}
    =
    \frac{1}{|\mathcal{B}|}
    \sum_{i\in\mathcal{B}}
    \mathbb{I}
    \left[
        \hat{y}^{\,\mathrm{fused}}_{i}\in\{1,\bot\}
    \right].
    \label{eq:bscer}
\end{equation}
where \(\bot\) denotes a pair for which no usable decision is returned; such a
pair is counted as an error for its ground-truth class. We report
\begin{equation}
    \mathrm{ACER}
    =
    \frac{\mathrm{MACER}+\mathrm{BSCER}}{2},
    \label{eq:ACER}
\end{equation}
together with EER computed from the fused morph risk scores. The repeated
response fusion rule, invalid response handling, reliability measures, and
security loss used for validation are defined in
Section~\ref{sec:fusion}.

\subsection{Prompt Generation and Selection Framework}
\label{sec:agents}

Figure~\ref{fig:dmad_pipeline} separates the framework into an offline
phase for prompt generation, validation, and selection and an online phase for
test evaluation. Version~1 sends generated candidates directly to validation
and stores the selected prompt in the Frozen Prompt Bank. Version~2 routes
aggregate validation metrics to the Prompt Optimizer Agent; its revised
candidates are screened by the Prompt Critic Agent before entering the next
Candidate Pool. The framework therefore contains three role-conditioned VLM
components---Prompt
Generator, Prompt Critic, and Prompt Optimizer---together with the
frozen-parameter
D-MAD Evaluator VLM and deterministic scoring and fusion modules.

\noindent\textbf{Prompt Generator Agent:}
The Prompt Generator Agent receives the task description, constraints, seed
prompt, and output schema of Fig.~\ref{fig:dmad_pipeline}. The constraints
prohibit protected attribute cues, require observable biometric and
image-forensic evidence, and enforce a structured JSON response. It produces
50 candidates for Version~1 and 30 Generation~0 candidates for Version~2 at
temperature \(0.8\). It never receives validation pairs, subject identities,
or labels.

\noindent\textbf{Prompt Critic Agent:}
The Prompt Critic Agent screens candidate text for protected attribute cues,
assumptions about a known morph generator, requests for hidden reasoning, unsafe
content, and omissions in the required JSON instruction. It is applied only to
Version~2 candidates from Generation~1 onwards; Generation~0 and the Version~1
pool are produced directly from the common task specification and bypass this
screen. The Critic Agent never receives facial images or class labels.

\noindent\textbf{D-MAD Evaluator VLM and Scoring Module:}
The D-MAD Evaluator VLM receives a candidate prompt and a validation image
pair \((A_i,B_i)\). No model parameters are updated, and only the prompt text
varies. Version~1 uses three responses per validation pair and applies the
fusion rule in Section~\ref{sec:fusion}; Version~2 uses one response per pair
during recursive optimisation, as described in Section~\ref{sec:v2}. The
Scoring Module converts the decisions and morph risk scores into MACER, BSCER,
ACER, reliability measures, and \(\mathcal{L}_{\mathrm{sec}}\). These
quantities are used only for ranking and optimisation on the validation split.

\noindent\textbf{Prompt Optimizer Agent:}
The Prompt Optimizer Agent is active only in Version~2. It receives selected
prompt texts, aggregate validation metrics, and a summary of the largest error
terms, but no test data, subject identities, image pairs, individual
predictions, or labels for individual pairs. It uses this feedback to produce
revised candidates targeting lower validation error, better response validity,
or greater decision stability. The optimiser is the same underlying model
under a different role instruction and emits revised candidates directly; no
separate Prompt Generator call is made during recursive revision.

\noindent\textbf{Frozen Prompt Bank and online evaluation:}
Prompts selected on the validation split are stored in the Frozen Prompt Bank before any test query. The online phase retrieves the predetermined Version~1 or Version~2 prompt and applies it to each test pair with the same evaluator, three responses per pair, and the fusion rules of Section~\ref{sec:fusion}.

\subsection{Repeated Response Fusion and Security Loss}
\label{sec:fusion}

The final evaluation uses three separate API calls for each image pair under
the same prompt, evaluator configuration, and inference temperature of zero.
Let \(\hat{y}_{i,r}\in\{0,1\}\) denote the binary decision for pair \(i\) in
response \(r\), where \(1\) denotes a morphing attack, and let
\(s_{i,r}\in[0,1]\) denote the corresponding morph risk score after rescaling
the returned value from \(0\)--\(100\). A response is decision-valid when it
can be parsed into the required schema and contains a usable binary decision;
the corresponding index set is \(\mathcal{V}^{(d)}_{i}\). The subset that also
contains a finite morph risk score is denoted by \(\mathcal{V}^{(s)}_{i}\).

Binary decisions are combined using logical OR:
\begin{equation}
\hat{y}^{\,\mathrm{fused}}_{i}
=
\begin{cases}
1, & \exists r\in\mathcal{V}^{(d)}_{i}\ \text{such that}\ \hat{y}_{i,r}=1,\\
0, & \mathcal{V}^{(d)}_{i}\neq\varnothing\ \text{and}\
      \hat{y}_{i,r}=0\ \forall r\in\mathcal{V}^{(d)}_{i},\\
\bot, & \mathcal{V}^{(d)}_{i}=\varnothing,
\end{cases}
\label{eq:decision_fusion}
\end{equation}
where \(\bot\) denotes an evaluator failure. This rule gives priority to attack indications across repeated responses and is therefore aligned with the objective of reducing missed morphing attacks. The corresponding score is obtained by maximum fusion:
\begin{equation}
S^{\,\mathrm{fused}}_{i}
=
\begin{cases}
\displaystyle\max_{r\in\mathcal{V}^{(s)}_{i}} s_{i,r},
    & \mathcal{V}^{(s)}_{i}\neq\varnothing,\\
\mathrm{NA}, & \mathcal{V}^{(s)}_{i}=\varnothing.
\end{cases}
\label{eq:score_fusion}
\end{equation}
The fused scores support ranking, EER computation, and threshold analysis; they are not interpreted as calibrated probabilities. Section~\ref{sec:repeat_ablation} examines the effect of using one, two, and three responses through an ablation based on the saved test outputs.

Invalid responses are assigned to two mutually exclusive categories. A parse
failure (PFR) occurs when non-empty returned text cannot be parsed into the
required JSON structure. A failure to answer (FTA) occurs when no response is
returned because of an empty output or API failure, or when a parseable response
contains a refusal or no usable binary morph decision. FTA and PFR are computed
as proportions of all requested responses. Fusion uses all decision-valid
responses. If all responses for a pair are invalid, the pair is counted as an
error for its ground-truth class, as formalised in Eqs.~\eqref{eq:macer}
and~\eqref{eq:bscer}. No fused score is assigned, and the pair is excluded from
score-based metrics. The inconsistency rate is the proportion of pairs with
conflicting decision-valid responses among pairs containing at least two such
responses.

Prompt ranking uses the following security loss:
\begin{equation}
\begin{split}
\mathcal{L}_{\mathrm{sec}}
=\;&
3\,\mathrm{MACER}
+\mathrm{BSCER}
+\mathrm{FTA}\\
&+0.5\,\mathrm{InconsistencyRate}
+0.5\,\mathrm{PFR},
\end{split}
\label{eq:lsec}
\end{equation}
where all terms are fractions in \([0,1]\) and lower values are better. The
coefficients are fixed design parameters rather than learned quantities: MACER
is weighted most heavily to reflect the security objective of reducing missed
attacks, BSCER penalises false rejection of bona fide pairs, FTA and PFR
penalise failures that prevent reliable operation, and the smaller
inconsistency coefficient discourages unstable responses without dominating
the classification errors. Their effect on the recursive Version~2 ranking is
examined in Section~\ref{sec:optimisation_analysis}.

Version~1 applies this security loss after repeated validation responses and decision fusion. Version~2 uses one evaluator response per validation pair during recursive optimisation to limit the number of validation queries. Its inconsistency term is therefore zero during prompt evolution, while repeated response reliability is measured during the final test evaluation. This distinction separates the prompt generation temperature and validation protocol from the test protocol with three responses per pair.

\subsection{Version 1: Non-Recursive Prompt Selection}
\label{sec:v1}

Version~1 examines whether prompt selection based on validation performance can improve a frozen-parameter VLM evaluator without recursive prompt revision. The Prompt Generator Agent produces 50 complete candidate prompts from a common D-MAD task description, output schema, and set of content constraints. The candidates differ in their emphasis on facial geometry, local identity
consistency, visible morphing artefacts, image quality, and final decision
formulation. During evaluation, every candidate is combined with the same evaluator model, system instruction, image role description, JSON schema, and inference settings. This design isolates the effect of the candidate prompt while keeping the remaining components fixed.

Each candidate is evaluated on the validation split with disjoint subjects. The validation labels are used to compute MACER, BSCER, ACER, \(\mathrm{FTA}\), \(\mathrm{PFR}\), and the inconsistency rate. These measurements guide prompt selection but do not update the VLM parameters or alter individual predictions. Version~1 should therefore be understood as prompt selection using validation labels rather than a label-free procedure.

For each candidate prompt \(p\), the selection procedure first computes the security loss defined in Eq.~\eqref{eq:lsec}. Candidates are then ordered using
\begin{equation}
    \mathbf{r}(p)
    =
    \left(
        \mathcal{L}_{\mathrm{sec}}(p),
        \mathrm{ACER}(p),
        \mathrm{MACER}(p)
    \right),
    \label{eq:v1_ranking}
\end{equation}
where lower values are preferred. The ranking is lexicographic: \(\mathcal{L}_{\mathrm{sec}}\) provides the primary criterion, ACER resolves equal security loss values, and MACER provides the final tie-breaking criterion. This rule fixes the selection procedure on the validation split and prevents performance on the test set from influencing the reported Version~1 system.

The five highest-ranked prompts are frozen after validation. Prompt
\texttt{P033} occupies rank one and defines the Version~1 test condition. The
remaining four are retained in the experiment records and do not influence the
reported Version~1 performance.

The manually designed prompt is evaluated using the same test pairs, VLM, repeated response protocol, and fusion rules. It does not participate in candidate generation or validation ranking and serves only as the reference condition for measuring the effect of automatic prompt selection.

\subsection{Version 2: Recursive Prompt Optimisation}
\label{sec:v2}

Version~2 examines whether aggregate validation feedback can improve D-MAD
prompts through successive revisions. The evaluator configuration is held
constant throughout this process, and only the prompt text changes. The Prompt Optimizer
Agent receives selected prompt texts, their aggregate validation metrics, and a
summary of the largest observed error terms. It does not receive test data,
subject identities, or validation outputs for individual pairs. Because the aggregate metrics are computed from validation labels, this stage
uses supervised validation feedback for prompt optimisation and is not a
label-free procedure.

Generation~0 contains 30 prompts produced by the Prompt Generator Agent. Each
prompt is evaluated once per validation pair with the evaluator temperature set
to zero. This single-response protocol limits the number of validation queries
required during recursive optimisation. It also means that repeated response
inconsistency is not informative at this stage. Reliability under repeated
queries is therefore assessed separately during final test evaluation. Prompt
generation and revision use a temperature of 0.8 to encourage variation,
whereas prompt evaluation and critic screening use a temperature of zero.
Repeated calls at temperature zero are treated as separate observations rather
than assumed to be deterministic.

After each generation, prompts are ranked using the validation value of
$\mathcal{L}_{\mathrm{sec}}$. The five prompts with the lowest security loss and
the three prompts with the lowest ACER are retained, with duplicate selections
removed. The Prompt Optimizer Agent receives the selected prompts, their
aggregate metrics, and a summary of the principal validation weaknesses across
the population. It then generates 15 revised prompts for the next generation. From
Generation~1 onwards, the Critic Agent screens these candidates before
evaluation. The optimiser consequently uses aggregate validation evidence without
receiving individual predictions or test information.

The recursive procedure allows at most three refinement generations beyond
Generation~0 and terminates when the best validation security loss fails to
improve for two consecutive generations. The executed experiment evaluated 30
prompts in Generation~0, 15 prompts in Generation~1, and 15 prompts in
Generation~2. The corresponding best values of
$\mathcal{L}_{\mathrm{sec}}$ were 1.10, 1.10, and 1.20. Generation~1 therefore
matched, but did not improve upon, the best initial prompt, while Generation~2
produced a higher minimum loss. The stopping condition was reached after
Generation~2, and Generation~3 was not evaluated.

For the revised-descendant comparison, \texttt{G1\_P014} is the reported
Version~2 prompt. It attains the minimum validation security loss of \(1.10\)
among the evaluated descendants and was fixed before test evaluation.
Generation~0 prompts are excluded only
from this descendant-specific condition because they have not undergone
revision; the best retained Generation~0 candidate is reported separately in
Section~\ref{sec:optimisation_analysis}. The final test evaluation applies the
protocol with three responses and the fusion rules of Section~\ref{sec:fusion}.

\section{Experiments and Results}
\label{sec:experiment}
This section describes the dataset, evaluation protocol, comparison settings, and implementation details used to assess the proposed prompt-selection framework.
\subsection{Dataset and Evaluation Protocol}
\label{sec:dataset}

We use the D-MAD dataset introduced in prior zero-shot VLM
work~\cite{Shekhawat2025}. Each sample contains a trusted bona fide
reference image \(A\) and a document image \(B\). The document image is
either genuine or generated using one of three morphing methods:
landmark-based morphing (LMA), MIPGAN-2, or PIPE.
The morphing method is
retained only as evaluation metadata and is not provided to the Prompt
Generator Agent, Prompt Critic Agent, Prompt Optimizer Agent, D-MAD
Evaluator VLM, or fusion procedure.

Following the 50-subject evaluation subset of~\cite{Shekhawat2025}, we use
50 subjects. Each selected subject
contributes one bona fide pair and one attack pair for each morphing method.
The subjects are divided into 10 for validation and 40 for testing, with no
identity shared between the two partitions. The validation split therefore
contains 10 bona fide and 30 attack pairs and is used only to evaluate
candidate prompts, compute \(\mathcal{L}_{\mathrm{sec}}\), and select the
prompts stored in the Frozen Prompt Bank. No test image, label, or metric
informs prompt generation, revision, ranking, or selection. The consequences
of this validation sample size are analysed in
Section~\ref{sec:optimisation_analysis}.

Evaluating a D-MAD prompt requires transmitting pairs of identifiable facial
images to an external multimodal service. The licences and consent conditions
of the other public benchmarks considered for this study did not authorise such
external processing. We therefore restricted the evaluation to the
collection for which participants had explicitly consented to the required
processing of their facial images. The one-dataset scope follows this privacy
and consent requirement rather than a convenience-based dataset choice.

The test partition therefore contains 160 image pairs: 40 bona fide,
40 LMA, 40 MIPGAN-2, and 40 PIPE. BSCER is computed over the 40 bona fide
pairs, overall MACER over the 120 attack pairs, and MACER for each morphing method over
40 pairs per method. All prompt systems are evaluated on these same pairs
under the fusion protocol of Section~\ref{sec:fusion}.

\subsection{Compared Methods and Implementation}
\label{sec:baselines}

We consider two reference settings. First, ViT~\cite{ViT},
DINOv2~\cite{DINOv2}, and CLIP~\cite{CLIP} serve as frozen visual
representation references. Each embeds Images~\(A\) and~\(B\) independently,
uses cosine distance as the comparison score, and applies a fixed
threshold to obtain a binary decision. Like the prompt systems, they involve
no model-parameter training or fine-tuning on morphing data. The prompt systems
nevertheless use labelled validation pairs for instruction selection.

Let \(\mathbf{z}_{A}\) and \(\mathbf{z}_{B}\) denote the embeddings of
Images~\(A\) and~\(B\). For each frozen representation, the comparison score
is the cosine distance
\(d(A,B)=1-\cos(\mathbf{z}_{A},\mathbf{z}_{B})\), and a pair is classified
as an attack when \(d(A,B)>0.2\). The threshold is fixed a priori and is held
constant across the three representations and all test pairs. Because cosine
distance distributions are model-dependent, these rows are interpreted only as
fixed-operating-point references rather than as calibrated or separately
optimised baselines. Cross-row differences at this threshold are therefore
descriptive and are not used to claim superiority over these representations.
They are not combined with the proposed roles or prompts and are not presented
as supervised D-MAD methods.

The selected instructions emphasize surface evidence rather than identity-level
comparison: \texttt{P033} directs the evaluator to ``unnatural transitions,
blending artefacts, and texture inconsistencies'' and \texttt{G1\_P014} to
``blending artefacts and skin texture uniformity''.

Second, a manually designed multi-step prompt provides the primary
controlled VLM baseline. It follows the design of~\cite{Shekhawat2025}, namely
the guided comparison of facial geometry, blending artefacts,
identity consistency and non-facial cues, but omits that work's introductory
role-conditioning statement. It is placed in the identical task wrapper used
by all generated candidates, which supplies the Image~\(A\)/Image~\(B\) role
definition and required JSON schema. It is evaluated with the same evaluator,
test pairs, temperature, and three-response fusion protocol. It differs from
the generated prompts only in the task instruction text. This comparison
isolates validation-guided prompt selection under a fixed protocol; it is not a
reproduction of the configuration published
in~\cite{Shekhawat2025}, whose results are computed over all 50 pairs per
morphing technique, including the subjects reserved here for validation, and
are therefore not directly comparable to the test results reported
below.

All queries were issued against the mutable \texttt{gpt-4o} alias rather than
a pinned snapshot. The exact model snapshot therefore cannot be reconstructed
from the saved outputs, and later evaluations using the same alias may use
different underlying weights. All prompt roles and the D-MAD evaluator use
GPT-4o, with the Generator and Optimizer roles at temperature \(0.8\) and the
Critic and evaluator roles at \(0\). Every candidate shares the same system
instruction, image ordering, task wrapper, and JSON schema, so only the
candidate prompt text changes; no model parameters are trained or updated by
the proposed procedure.

The manual prompt, \texttt{P033}, and \texttt{G1\_P014} are each evaluated three times per test
pair, and all reported metrics are computed from the saved structured outputs
after the fusion of Section~\ref{sec:fusion}, with no further VLM queries and
without prompt selection on the test set. To support reproduction we will release the
candidate prompts of both versions, the ranking and freezing scripts, and the
saved structured outputs, excluding all facial images, image paths, and subject
identifiers.

\subsection{Overall D-MAD Performance}
\label{sec:main_results}

Table~\ref{tab:main_comparison} reports overall performance for the
prompts selected on the validation split, the manual prompt, and the frozen visual
references. Among the prompt conditions, \texttt{P033} attains the lowest
error, with ACER \(=9.58\%\), MACER \(=19.17\%\), and
EER \(=6.67\%\), an observed MACER difference of \(50.83\) percentage
points relative to the manual prompt. Both conditions share the same evaluator
configuration, test pairs, task wrapper, response schema, and fusion rule; the
instruction condition is the controlled experimental variable.

All three prompt conditions yield \(0.00\%\) BSCER after fusion of three
responses, so ACER equals MACER/2. Two \texttt{P033} responses on bona fide
pairs are parse failures, but both pairs retain valid responses and are
accepted after fusion. The zero observed BSCER on 40 pairs does not rule out
false rejections in deployment; thresholding the uncalibrated fused scores
instead provides a score level BSCER--MACER trade-off.

\begin{table}[!tb]
\centering
\caption{Overall test performance on 40 bona fide and 120 attack pairs.
Version~1 and Version~2 prompts are selected using labelled validation pairs;
no model parameters are updated. Lower values are better.}
\label{tab:main_comparison}
\setlength{\tabcolsep}{3.4pt}
\renewcommand{\arraystretch}{1.05}
\resizebox{\columnwidth}{!}{%
\begin{tabular}{lcccc}
\toprule
\textbf{Method} &
\textbf{HTER} &
\textbf{MACER} &
\textbf{BSCER} &
\textbf{EER} \\
& \textbf{(\%)} & \textbf{(\%)} & \textbf{(\%)} & \textbf{(\%)} \\
\midrule
ViT~\cite{ViT}                 & 12.50 & 25.00 & 0.00 & 9.79 \\
DINOv2~\cite{DINOv2}           & 42.08 & 84.17 & 0.00 & 15.42 \\
CLIP~\cite{CLIP}               & 46.25 & 92.50 & 0.00 & 22.50 \\
Manual prompt (adapted from~\cite{Shekhawat2025})
                               & 35.00 & 70.00 & 0.00 & 10.83 \\
V1: \texttt{P033}       & \textbf{9.58} &
                                 \textbf{19.17} &
                                 0.00 &
                                 \textbf{6.67} \\
V2: \texttt{G1\_P014}                   & 15.83 & 31.67 & 0.00 & 7.50 \\
\bottomrule
\end{tabular}%
}
\end{table}

The recursively generated \texttt{G1\_P014} prompt improves on the manual
prompt but does not surpass \texttt{P033}, reaching \(31.67\%\) MACER.
Because the frozen visual references use a common uncalibrated threshold, their
decision-level rows provide context rather than a basis for an algorithmic
superiority claim. Statistical comparisons between prompt conditions appear in
Section~\ref{sec:statistics}.

\subsection{Morph Specific Analysis}
\label{sec:morph_specific}

Table~\ref{tab:morph_specific} separates MACER by morphing method. \texttt{P033}
detects all 40 MIPGAN-2 and all 40 PIPE attacks but misses 23 of the 40 LMA
attacks, so its overall MACER of \(19.17\%\) is determined entirely by
landmark-based morphing. \texttt{G1\_P014} follows the same pattern with 33 missed
LMA and five missed PIPE attacks, while the manual prompt errs heavily on
all three methods.

\begin{table}[!tb]
\centering
\caption{Test MACER by morphing method, with missed attacks among 40 pairs in
parentheses. Lower is better.}
\label{tab:morph_specific}
\setlength{\tabcolsep}{4.0pt}
\renewcommand{\arraystretch}{1.05}
\resizebox{\columnwidth}{!}{%
\begin{tabular}{lccc}
\toprule
\textbf{Method} &
\textbf{LMA} &
\textbf{MIPGAN-2} &
\textbf{PIPE} \\
& \textbf{MACER (\%)} &
  \textbf{MACER (\%)} &
  \textbf{MACER (\%)} \\
\midrule
Manual prompt
    & 85.00 (34) & 45.00 (18) & 80.00 (32) \\
V1: \texttt{P033}
    & \textbf{57.50 (23)} & \textbf{0.00 (0)} & \textbf{0.00 (0)} \\
V2: \texttt{G1\_P014}
    & 82.50 (33) & \textbf{0.00 (0)} & 12.50 (5) \\
\bottomrule
\end{tabular}%
}
\end{table}

The selected prompts achieve zero observed missed attacks on the tested
MIPGAN-2 subsets and, for \texttt{P033}, on the PIPE subset, whereas LMA
remains substantially more difficult. The zero observed error
for a morphing method is limited to the 40 test pairs considered here and
should not be interpreted as evidence of perfect detection beyond this
evaluation.

\subsection{Response Budget and Reliability}
\label{sec:repeat_ablation}

We evaluate the response budget using the three saved test responses per
pair. For one or two responses, results are averaged over all subsets of
that size; the three-response condition uses all outputs. Each subset
applies the OR decision fusion of Section~\ref{sec:fusion}.
Table~\ref{tab:repeat_ablation} reports the resulting MACER together with
the reliability rates discussed below. This analysis reuses the fixed saved
calls and is not an independently repeated experiment.

\begin{table}[htp]
\centering
\caption{Test response budget ablation. MACER averages all subsets; FTA and
PFR are per response rates, and inconsistency is measured per eligible pair.}
\label{tab:repeat_ablation}
\setlength{\tabcolsep}{3.0pt}
\renewcommand{\arraystretch}{1.05}
\resizebox{\columnwidth}{!}{%
\begin{tabular}{lcccccc}
\toprule
& \multicolumn{3}{c}{\textbf{MACER (\%)}} &
  \multicolumn{3}{c}{\textbf{Reliability (\%)}} \\
\cmidrule(lr){2-4}\cmidrule(lr){5-7}
\textbf{Method} & \textbf{1 resp.} & \textbf{2 resp.} & \textbf{3 resp.} &
  \textbf{FTA} & \textbf{PFR} & \textbf{Incons.} \\
\midrule
Manual prompt & 72.50 & 71.11 & 70.00 & 0.00 & 0.00 & 3.13 \\
V1: \texttt{P033}      & 23.33 & 20.00 & \textbf{19.17} & 0.00 & 0.63 & 6.88 \\
V2: \texttt{G1\_P014}  & 32.78 & 31.94 & 31.67 & 0.00 & 0.00 & \textbf{1.88} \\
\bottomrule
\end{tabular}%
}
\end{table}

Across the saved calls, OR fusion reduces \texttt{P033} MACER from
\(23.33\%\) with one response to \(19.17\%\) with three, with most of the
gain obtained at two responses. The corresponding reductions for the manual
and Version~2 prompts are \(2.50\) and \(1.11\) percentage points. The one
response \texttt{P033} condition has \(1.67\%\) average BSCER because two
bona fide responses are parse failures; fusion recovers all affected pairs
when multiple responses are available.
 \texttt{P033} produces three parse failures among 480
responses (\(0.63\%\)) and has the highest inconsistency rate
(\(6.88\%\), compared with \(3.13\%\) and \(1.88\%\)). Repeated calls
therefore improve attack sensitivity.

\subsection{Statistical Analysis}
\label{sec:statistics}

Each subject contributes multiple test pairs, so observations for pairs
are not independent. We therefore use a nonparametric bootstrap over the
40 test subjects as the primary uncertainty analysis. Subjects are sampled
with replacement for \(100{,}000\) iterations, and MACER is recomputed
from all attack pairs associated with each sampled subject. Relative to
\texttt{P033}, the manual prompt increases MACER by \(50.83\) percentage points,
with a \(95\%\) percentile interval of
\([43.33,58.33]\). \texttt{G1\_P014} increases MACER by \(12.50\) percentage
points relative to \texttt{P033}, with an interval of \([6.67,18.33]\).
At the subject level, \texttt{P033} has fewer missed attacks than the manual prompt
for 37 of 40 subjects and ties it for the remaining three. It has fewer missed
attacks than \texttt{G1\_P014} for 13 subjects and ties it for 27.

Because pairs from the same subject are dependent, we do not interpret
pair-level significance tests as the primary inferential analysis. The paired
subject bootstrap and the subject-level win/tie counts above provide the
uncertainty evidence used for the prompt comparisons.

\subsection{Prompt Optimisation Analysis}
\label{sec:optimisation_analysis}

The best validation security losses were \(1.10\), \(1.10\), and \(1.20\)
for Generations~0, 1, and 2, respectively. Generation~1 therefore matched, but
did not improve upon, the best Generation~0 prompt, while Generation~2 produced
a higher minimum loss.

With 30 validation attacks, one additional miss changes MACER by \(3.33\)
percentage points and, when the other loss terms are zero, changes
\(\mathcal{L}_{\mathrm{sec}}\) by \(0.10\). The difference between the best
Generation~0 and Generation~2 losses therefore corresponds to one attack
decision. Among the surviving lineages, the weakest ancestor improves from
\(1.50\) to \(1.10\), whereas the strongest degrades from \(1.10\) to
\(1.20\). On the test split, the retained Generation~0 seed obtains slightly
lower post hoc MACER than the revised descendant (\(30.83\%\) versus
\(31.67\%\)); this comparison did not influence selection. Recursive revision
therefore does not improve the best candidate under the available validation
resolution, but this result does not preclude a benefit at a larger scale.

For the retained Version~2 candidates, BSCER, FTA, PFR, and inconsistency are
zero, so \(\mathcal{L}_{\mathrm{sec}}=3\,\mathrm{MACER}\). Consequently,
all tested positive MACER weights, equal weighting, and ranking by MACER alone
produce the same ordering (Kendall \(\tau=1.00\)). The additional loss terms
are therefore inactive for this validation population.


\section{Limitations and Future Work}
\label{sec:limitations}

The study uses one VLM configuration without parameter updates and one dataset
with disjoint validation and test subjects. This scope follows the consent
constraints of Section~\ref{sec:dataset}; the other considered benchmarks did
not authorise external processing of identifiable facial images. Transfer
across models, datasets, demographics, morph generators, and acquisition
conditions is therefore untested. The validation split of 10 subjects may
overfit prompt selection, while the mutable API alias limits temporal
reproducibility. No ablation removes or mismatches the trusted reference, so
the use of differential rather than document image evidence remains
unquantified. The visual references use a common uncalibrated threshold, and
the VLM risk scores are also uncalibrated. For consequential D-MAD decisions,
VLM outputs should only assist trained human review. External processing must
follow applicable consent, access, retention, and jurisdictional requirements.
We redistribute no images or subject identifiers.

\section{Conclusion}
\label{sec:conclusion}

The Version~1 prompt selected on validation data obtains \(19.17\%\) MACER,
compared with \(70.00\%\) for the manual prompt, at \(0.00\%\) observed
BSCER. It has fewer missed attacks for 37 of 40 test subjects, with no reverse
outcome. Recursive optimisation does not improve the best Generation~0
validation loss; its reported descendant obtains \(31.67\%\) test MACER.
Errors remain concentrated in LMA. These results show that prompt formulation
can have a substantial effect on the behaviour of a frozen VLM in D-MAD and
should therefore be treated as part of the evaluation protocol rather than as
a minor implementation choice. At the same time, the remaining LMA errors
suggest that relying mainly on visible blending and texture cues is not enough
for more challenging morphs, and that stronger use of differential identity
evidence may be needed. The findings are limited to the model and dataset
studied here and should not be interpreted as general superiority across
D-MAD systems. Future work should examine reference image use and transfer
across models and datasets while retaining human oversight.
{
    \small
    \bibliographystyle{ieeenat_fullname}
    \bibliography{FaceMorph_Refereces-4}
}

\end{document}


\maketitle

\section{Overview}
\label{supp:overview}

This supplement provides the material omitted from the main paper to satisfy
its page limit. It expands the experimental protocol, reproduces the selected
Version~1 and Version~2 task instructions, presents fused morph risk score
distributions, and gives representative qualitative outputs. It also reports
the response budget analysis, subject level uncertainty, recursive prompt
optimisation diagnostics, and the scope of the planned reproducibility
release. The numerical values in this document follow the same protocol as
the main paper: decisions are fused once per image pair, failures are
classified into mutually exclusive categories, and prompt selection uses only
the validation split.

\section{Extended Experimental Protocol}
\label{supp:protocol}

\subsection{Dataset composition and split}

The experiments use 50 subjects from the public D-MAD dataset introduced in
prior zero-shot VLM work~\cite{Shekhawat2025}. Each selected subject contributes one
bona fide comparison and one attack comparison for each of three morphing
methods: landmark based morphing (LMA), MIPGAN-2, and PIPE. Ten subjects form
the validation split, giving 10 bona fide and 30 attack pairs. The remaining
40 subjects form the test split, giving 40 bona fide, 40 LMA, 40 MIPGAN-2,
and 40 PIPE pairs. The split is disjoint by subject. Validation data are used
for prompt ranking and recursive revision, whereas the 160 test pairs are
used only after the selected prompts have been frozen.

The morphing method is retained as evaluation metadata but is not supplied to
the Prompt Generator Agent, Prompt Critic Agent, Prompt Optimizer Agent, the
D-MAD Evaluator VLM, or the fusion procedure. Identifiable facial images,
image paths, and subject identifiers are excluded from the planned release.

\subsection{Evaluator output schema}

All reported prompt conditions use the same image ordering, system
instruction, task wrapper, and structured response schema. Image~$A$ is the
trusted bona fide reference or live capture, and Image~$B$ is the document
image under evaluation. The experimental scripts retain the historical JSON
field name \texttt{morph\_probability}; throughout the paper and supplement,
this value is called a \emph{morph risk score} because it is not a calibrated
probability.

\begin{lstlisting}
{
  "same_identity": "yes/no/uncertain",
  "identity_confidence": 0-100,
  "morph_detected": "yes/no",
  "morph_probability": 0-100,
  "main_evidence": [
    "observable evidence 1",
    "observable evidence 2",
    "observable evidence 3"
  ],
  "quality_caveat": "short note",
  "decision_reliability": "low/medium/high"
}
\end{lstlisting}

The natural language evidence is retained as an audit rationale for the
returned decision. It is not treated as a faithful causal explanation of the
internal computation of the VLM.

\subsection{Repeated response fusion and invalid outputs}

For the final test evaluation, each prompt is queried three times per image
pair at evaluator temperature zero. Let $\mathcal{V}^{(d)}_{i}$ denote the
decision-valid responses for pair $i$, and $\mathcal{V}^{(s)}_{i}\subseteq
\mathcal{V}^{(d)}_{i}$ those also containing a finite morph risk score
$s_{i,r}\in[0,1]$. The pair level decision is
\begin{equation}
\hat{y}^{\mathrm{fused}}_{i}=
\begin{cases}
1, & \exists r\in\mathcal{V}^{(d)}_{i}:\hat{y}_{i,r}=1,\\
0, & \mathcal{V}^{(d)}_{i}\neq\varnothing\ \text{and}\
     \hat{y}_{i,r}=0\ \forall r\in\mathcal{V}^{(d)}_{i},\\
\bot, & \mathcal{V}^{(d)}_{i}=\varnothing,
\end{cases}
\label{eq:supp_decision_fusion}
\end{equation}
and the fused score is
\begin{equation}
S^{\mathrm{fused}}_{i}=\max_{r\in\mathcal{V}^{(s)}_{i}}s_{i,r},
\qquad \mathcal{V}^{(s)}_{i}\neq\varnothing.
\label{eq:supp_score_fusion}
\end{equation}
No fused score is assigned when $\mathcal{V}^{(s)}_{i}=\varnothing$. The fused
scores are used for ranking, threshold analysis, and EER, but are not
interpreted as calibrated probabilities.

A parse failure rate (PFR) event is a non-empty response that cannot be parsed into the required JSON structure. A failure to answer
(FTA) event is an empty output, an API failure, a refusal, or a parseable
response without a usable binary decision. These categories are mutually
exclusive. Decision fusion uses all decision-valid responses. If no such
response exists, the pair is treated as an evaluator failure and charged as an
error to its ground-truth class; it is excluded from score based measures when
no fused score exists. Decision inconsistency is computed only for pairs with
at least two decision-valid responses and records whether they disagree.

\subsection{Security loss and prompt selection}

Prompt ranking uses
\begin{equation}
\begin{split}
\mathcal{L}_{\mathrm{sec}}={}&3\,\mathrm{MACER}
+\mathrm{BSCER}+\mathrm{FTA}\\
&+0.5\,\mathrm{InconsistencyRate}+0.5\,\mathrm{PFR},
\end{split}
\label{eq:supp_security_loss}
\end{equation}
where every term is represented as a fraction in $[0,1]$. The coefficients
are fixed design parameters. They express an operating preference that places
the largest penalty on missed attacks while retaining penalties for bona fide
errors, invalid outputs, and unstable decisions.

Version~1 generates 50 candidates from a common task specification. Each
candidate is evaluated on the validation split using three responses per pair.
Candidates are ordered lexicographically by
$(\mathcal{L}_{\mathrm{sec}},\mathrm{ACER},\mathrm{MACER})$. The five highest
ranked candidates are frozen before test evaluation, and the rank-one prompt
\texttt{P033} defines the primary Version~1 condition.

Version~2 begins with 30 Generation~0 prompts and uses one evaluator response
per validation pair during recursive optimisation. After each generation, the
five prompts with the lowest security loss and the three with the lowest ACER
are retained, with duplicate selections removed. The Prompt Optimizer Agent
produces 15 descendants for the next generation, and the Prompt Critic Agent
screens candidates from Generation~1 onwards. The executed experiment
contains Generations~0, 1, and 2. Prompt \texttt{G1\_P014} is the highest
ranked recursively generated descendant on the validation split and defines
the reported Version~2 condition. No test metric contributes to either
selection.

\section{Selected Prompt Texts}
\label{supp:prompts}

The task instructions below are reproduced from the prompt records used to
construct the qualitative figures. Both are inserted into the same fixed
wrapper and evaluated with the schema of Section~\ref{supp:protocol}.

\subsection{Version~1 prompt \texttt{P033}}

\begin{quote}
\small
Conduct a comprehensive review of artefacts in Image~B compared with Image~A.
Focus on unnatural transitions, blending artefacts, and texture
inconsistencies. Return a structured decision with identity confidence, morph
decision, morph risk score, evidence, quality caveat, and reliability.
\end{quote}

\subsection{Version~2 prompt \texttt{G1\_P014}}

\begin{quote}
\small
Assess Image~A and Image~B for identity by focusing on blending artefacts and
skin texture uniformity. Avoid demographic considerations. Return JSON fields
for identity decision, confidence, morph decision, morph risk score, evidence,
quality caveat, and reliability.
\end{quote}

\subsection{Common fixed wrapper}

\begin{quote}
\small
Image~A is the trusted bona fide reference or live capture image. Image~B is
the document or passport style image to evaluate. Determine whether Image~B
appears to be a morphing attack relative to Image~A. The morph generation
method is unknown. Do not assume a specific generator or morphing technique.
Return only valid JSON and no text outside JSON. Do not reveal hidden chain of
thought. Use concise observable evidence only and follow the required response
schema.
\end{quote}

The manually designed baseline follows the multi-step comparison structure of
the preceding D-MAD study~\cite{Shekhawat2025}, covering facial geometry,
identity consistency, blending artefacts, and non-facial cues. In the current
experiment, that task instruction is placed in the same fixed wrapper and
response schema as the generated prompts. Its exact adapted text will be
released with the reproducibility material after the anonymous review process;
the published results of the preceding study are not reused because its
evaluation partition differs from the present test split.

\section{Fused Morph Risk Score Distributions}
\label{supp:distributions}

\begin{figure*}[t]
    \centering
    \begin{subfigure}[t]{.30\textwidth}
        \centering
        \includegraphics[width=\textwidth]{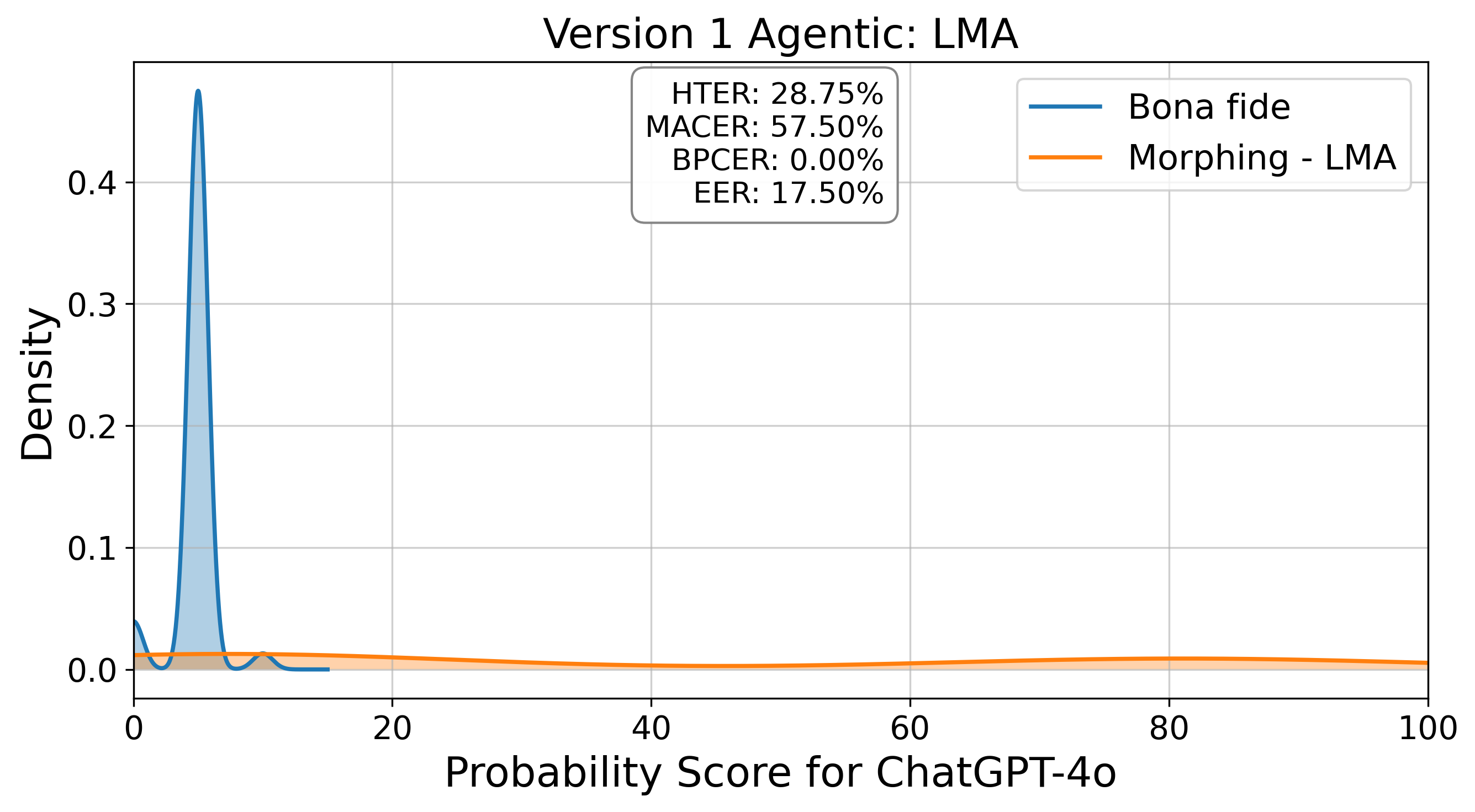}
        \caption{\texttt{P033}: LMA}
        \label{fig:supp_dist_v1_lma}
    \end{subfigure}
    \hspace{0.02\textwidth}
    \begin{subfigure}[t]{.30\textwidth}
        \centering
        \includegraphics[width=\textwidth]{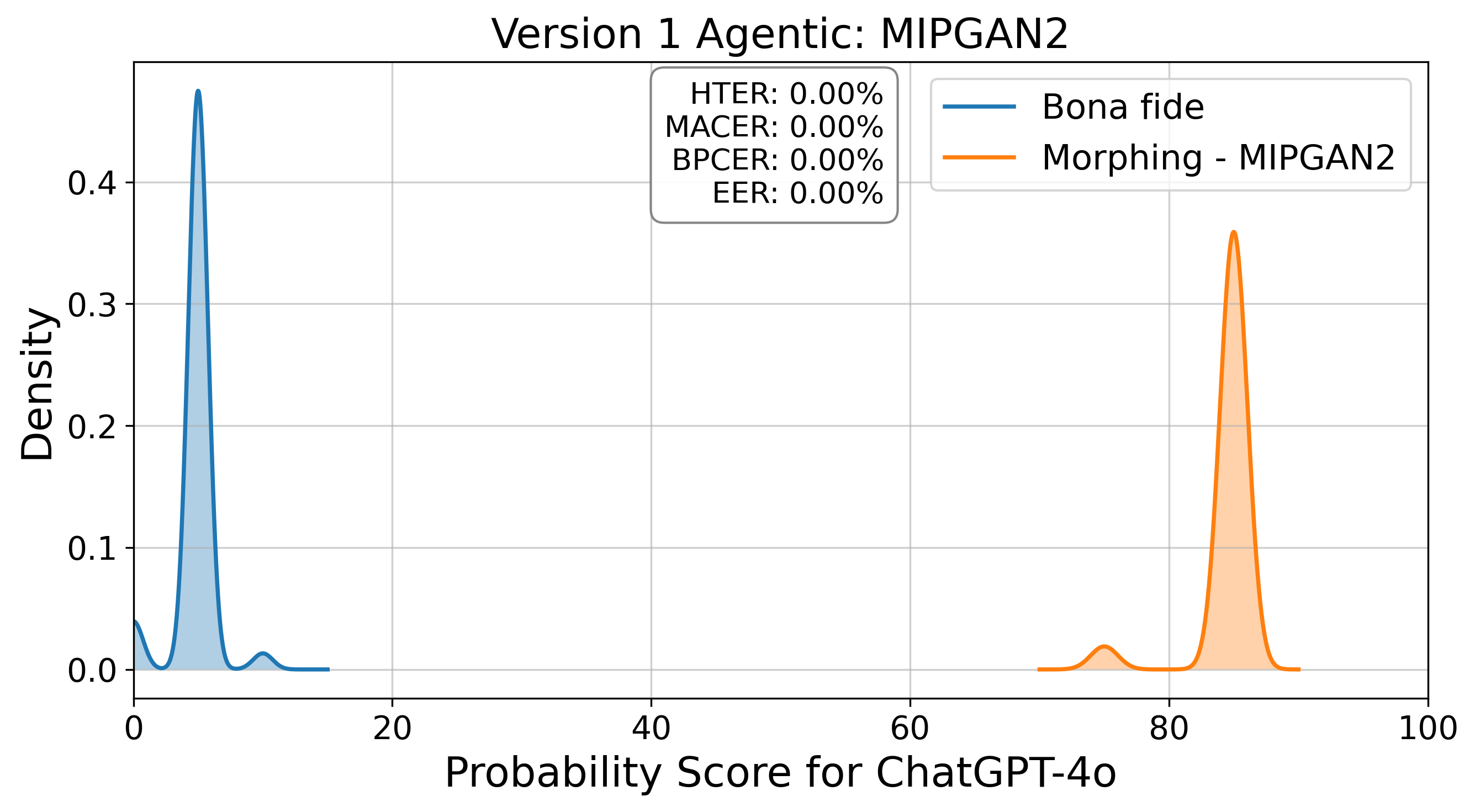}
        \caption{\texttt{P033}: MIPGAN-2}
        \label{fig:supp_dist_v1_mipgan}
    \end{subfigure}
    \hspace{0.02\textwidth}
    \begin{subfigure}[t]{.30\textwidth}
        \centering
        \includegraphics[width=\textwidth]{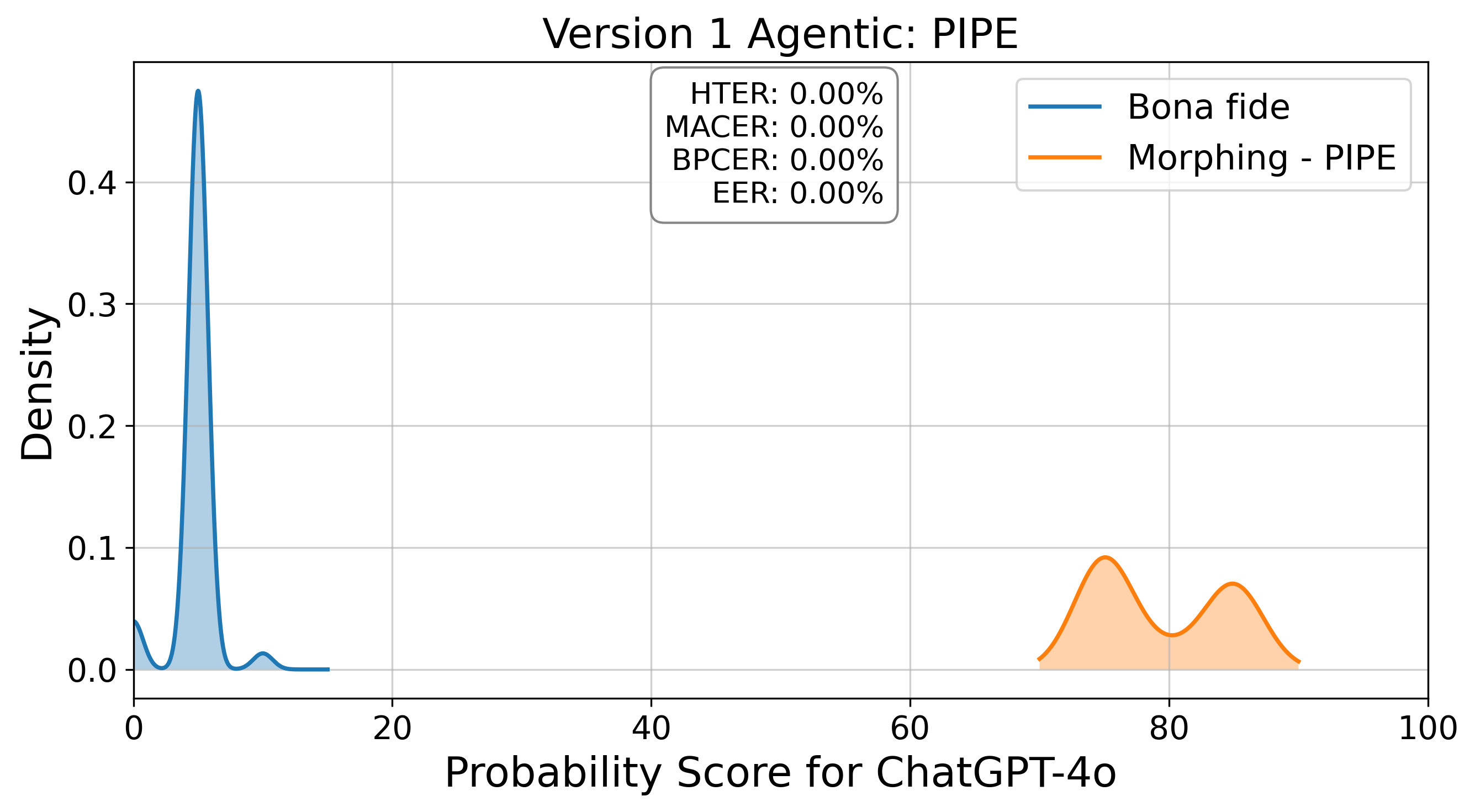}
        \caption{\texttt{P033}: PIPE}
        \label{fig:supp_dist_v1_pipe}
    \end{subfigure}\\
    \begin{subfigure}[t]{.30\textwidth}
        \centering
        \includegraphics[width=\textwidth]{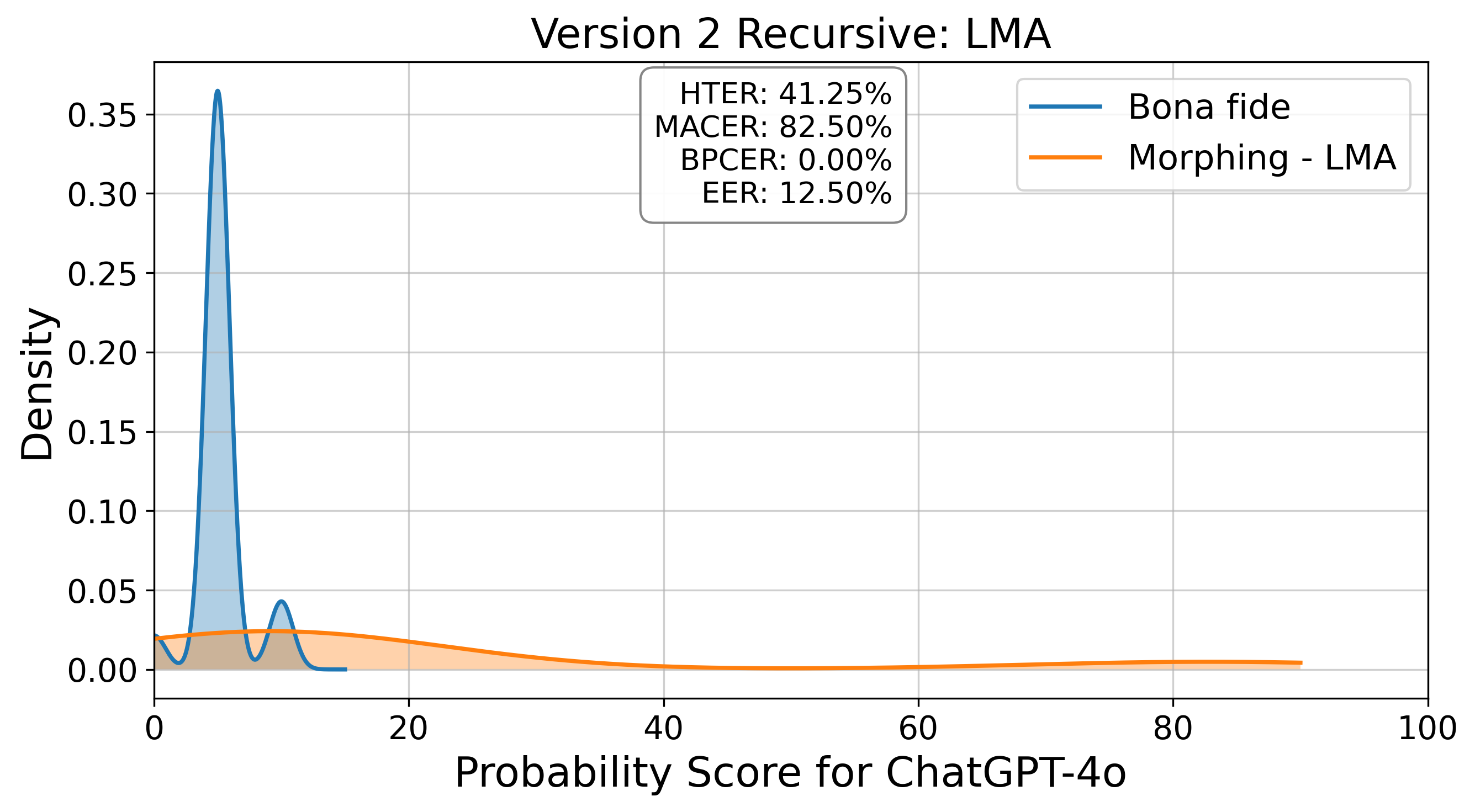}
        \caption{\texttt{G1\_P014}: LMA}
        \label{fig:supp_dist_v2_lma}
    \end{subfigure}
    \hspace{0.02\textwidth}
    \begin{subfigure}[t]{.30\textwidth}
        \centering
        \includegraphics[width=\textwidth]{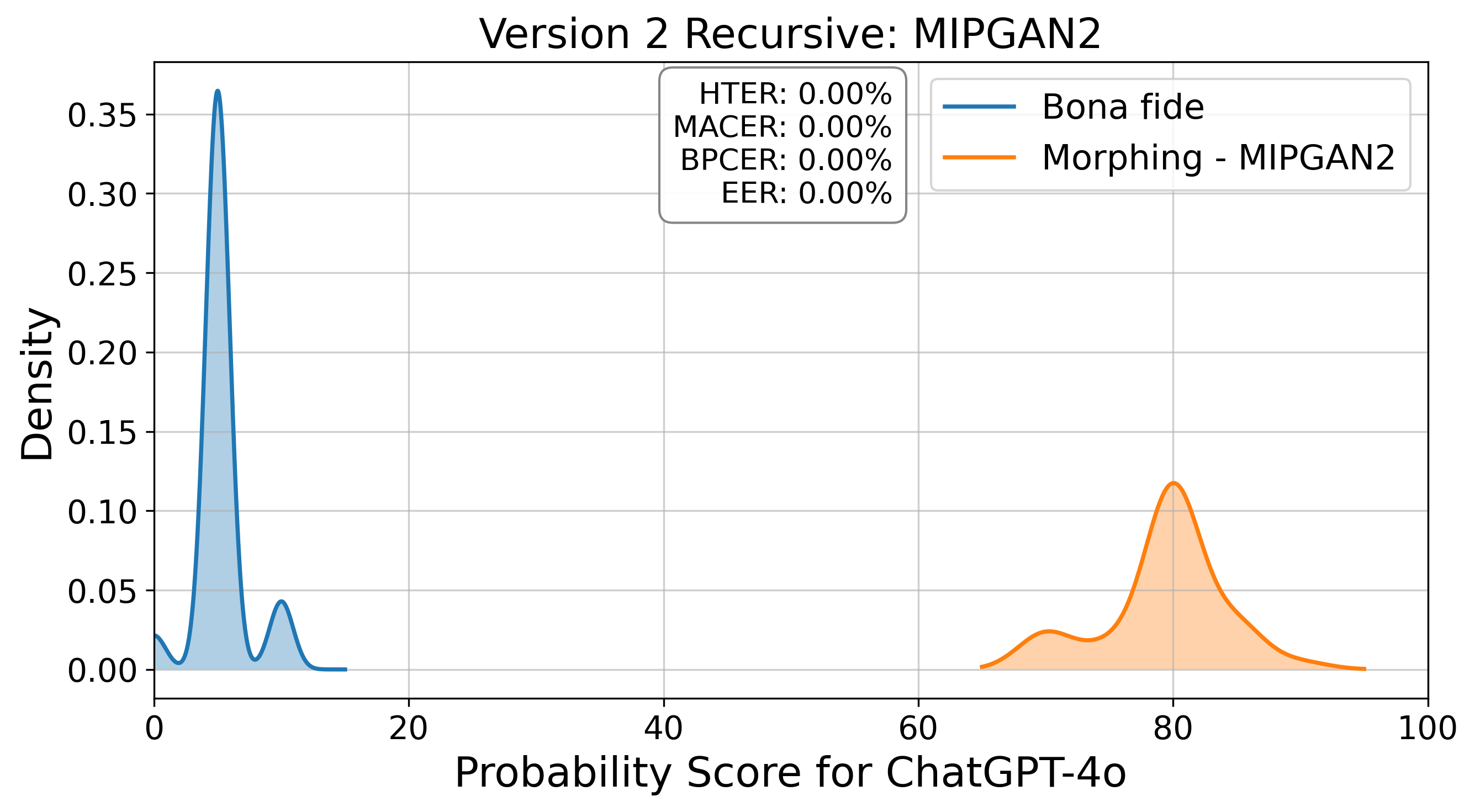}
        \caption{\texttt{G1\_P014}: MIPGAN-2}
        \label{fig:supp_dist_v2_mipgan}
    \end{subfigure}
    \hspace{0.02\textwidth}
    \begin{subfigure}[t]{.30\textwidth}
        \centering
        \includegraphics[width=\textwidth]{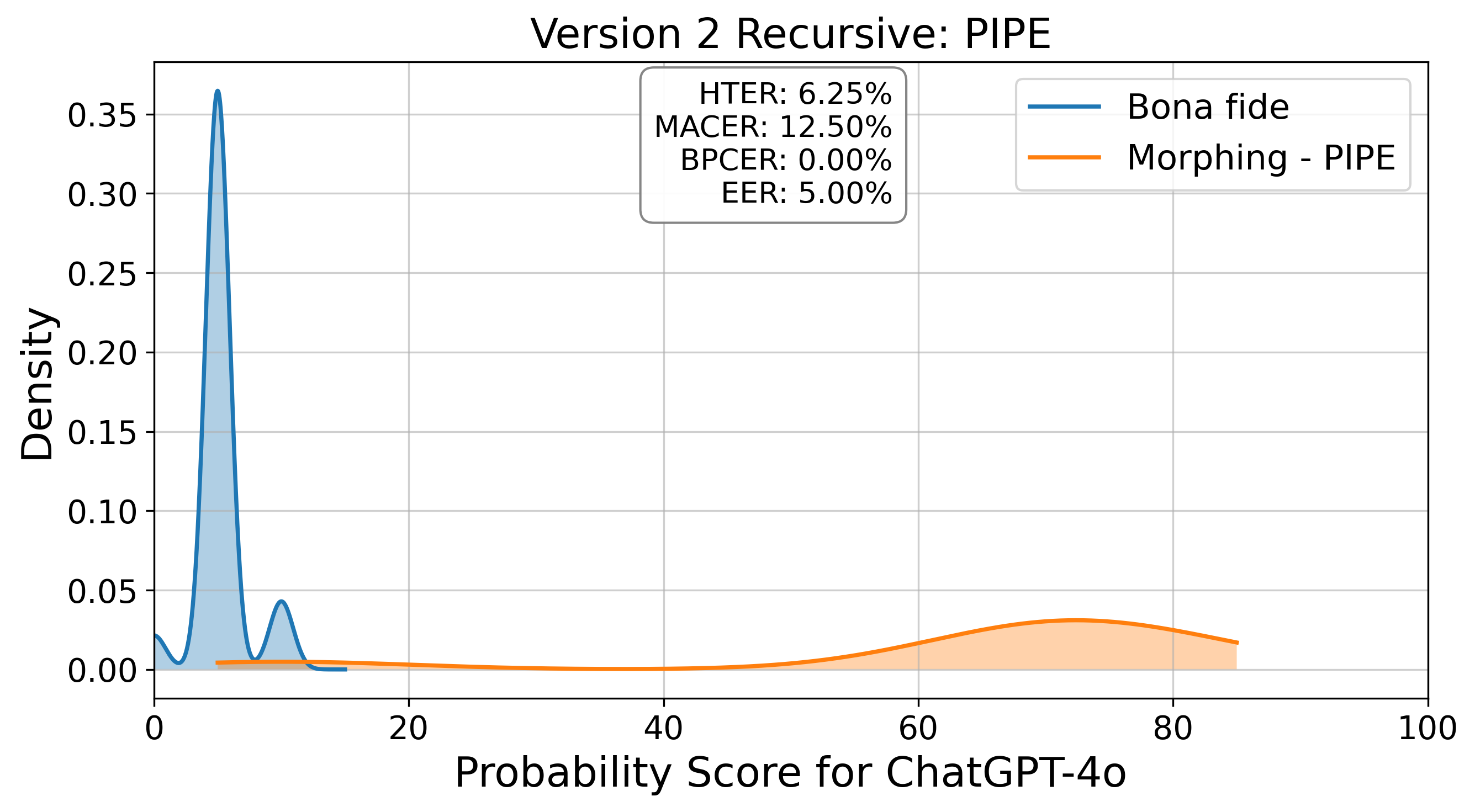}
        \caption{\texttt{G1\_P014}: PIPE}
        \label{fig:supp_dist_v2_pipe}
    \end{subfigure}
    \caption{Kernel density estimates of fused morph risk scores for bona fide
    and morphed comparisons. The top row corresponds to the selected Version~1
    prompt \texttt{P033}; the bottom row corresponds to the selected recursive
    descendant \texttt{G1\_P014}. Greater overlap between the bona fide and
    attack distributions indicates weaker score separation. The scores are
    uncalibrated and are not probabilities; historical plot labels are mapped
    to the manuscript terminology in the accompanying text.}
    \label{fig:supp_distribution}
\end{figure*}

Figure~\ref{fig:supp_distribution} presents kernel density estimates of the
fused morph risk scores for the two selected prompts. The plots use one fused
score per pair after maximum fusion over valid responses. The horizontal axis
must be interpreted as an uncalibrated morph risk score rather than a
probability. The embedded plots retain labels from the original analysis
script: ``Probability Score'' denotes the fused morph risk score, ``BPCER''
denotes BSCER, and the ``Agentic'' run label denotes the Version~1 condition.

The score distributions explain the error pattern reported in the main paper.
For \texttt{P033}, MIPGAN-2 and PIPE scores are separated from the bona fide
region, consistent with zero missed attacks for these two methods. The LMA
distribution overlaps substantially with the bona fide scores, explaining the
23 missed LMA attacks. The recursive prompt exhibits the same general pattern
but has greater overlap for LMA and a broader PIPE distribution, consistent
with 33 missed LMA attacks and five missed PIPE attacks.

\begin{table}[t]
\centering
\caption{MACER by morphing method on the 40 attack pairs of each type. Values
in parentheses are missed attacks. Lower is better.}
\label{tab:supp_morph_specific}
\setlength{\tabcolsep}{4.2pt}
\renewcommand{\arraystretch}{1.05}
\begin{tabular}{lccc}
\toprule
\textbf{Prompt} & \textbf{LMA} & \textbf{MIPGAN-2} & \textbf{PIPE} \\
\midrule
Manual & 85.00 (34) & 45.00 (18) & 80.00 (32) \\
\texttt{P033} & 57.50 (23) & 0.00 (0) & 0.00 (0) \\
\texttt{G1\_P014} & 82.50 (33) & 0.00 (0) & 12.50 (5) \\
\bottomrule
\end{tabular}
\end{table}

The 23 LMA pairs missed by \texttt{P033} form a strict subset of the 33 LMA
pairs missed by \texttt{G1\_P014}. This nested error pattern indicates that
the recursive prompt does not recover an alternative group of difficult LMA
pairs; instead, it misses all LMA cases missed by Version~1 and ten additional
cases.

\FloatBarrier
\section{Qualitative Analysis}
\label{supp:qualitative}

The qualitative examples are drawn from the held-out test partition. The
applicable consent and publication conditions authorise reproduction of these
examples. They
illustrate how the selected prompts describe a correct bona fide decision and
a false negative on an LMA pair. The returned text is presented as an audit
rationale and not as a causal explanation of the model's internal reasoning.
No subject identifiers or image paths are reported.

\begin{figure*}[t]
\centering
\scriptsize
\setlength{\tabcolsep}{3pt}
\renewcommand{\arraystretch}{1.02}

\begin{tabularx}{\textwidth}{@{}X@{}}
\toprule
\textbf{Selected Version~1 task instruction (\texttt{P033}).}
Conduct a comprehensive review of artefacts in Image~B compared with Image~A.
Focus on unnatural transitions, blending artefacts, and texture
inconsistencies. Return a structured decision with identity confidence, morph
decision, morph risk score, evidence, quality caveat, and reliability.\\
\bottomrule
\end{tabularx}

\vspace{1.5mm}
\begin{minipage}[t]{0.48\textwidth}
\centering
\includegraphics[width=0.62\linewidth]{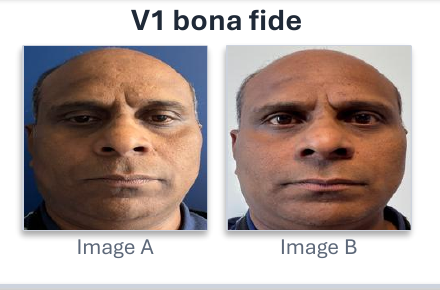}

\raggedright
\textbf{Bona fide, correct.} Fused morph risk score: $5/100$.
\textbf{Audit rationale:} The response reports consistent facial geometry and
landmarks around the eyes, nose, and mouth, with no visible blending,
ghosting, unnatural smoothing, or boundary artefacts. The reported reliability
is high.
\end{minipage}
\hfill
\begin{minipage}[t]{0.48\textwidth}
\centering
\includegraphics[width=0.62\linewidth]{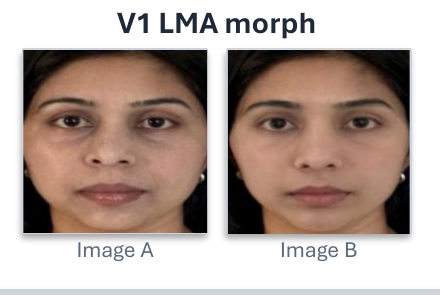}

\raggedright
\textbf{LMA, false negative.} Fused morph risk score: $5/100$.
\textbf{Audit rationale:} The response identifies no clear texture
discontinuity, local averaging, or boundary artefact and therefore predicts
bona fide. The example illustrates how an LMA image can preserve a natural
surface appearance and provide few visible artefacts for this prompt.
\end{minipage}

\caption{Representative outputs for the selected Version~1 prompt
\texttt{P033}. The bona fide pair is accepted correctly, whereas the LMA pair
is missed because the response identifies little visible blending or texture
evidence.}
\label{fig:supp_qualitative_v1}
\end{figure*}

\begin{figure*}[t]
\centering
\scriptsize
\setlength{\tabcolsep}{3pt}
\renewcommand{\arraystretch}{1.02}

\begin{tabularx}{\textwidth}{@{}X@{}}
\toprule
\textbf{Selected Version~2 task instruction (\texttt{G1\_P014}).}
Assess Image~A and Image~B for identity by focusing on blending artefacts and
skin texture uniformity. Avoid demographic considerations. Return JSON fields
for identity decision, confidence, morph decision, morph risk score, evidence,
quality caveat, and reliability.\\
\bottomrule
\end{tabularx}

\vspace{1.5mm}
\begin{minipage}[t]{0.48\textwidth}
\centering
\includegraphics[width=0.62\linewidth]{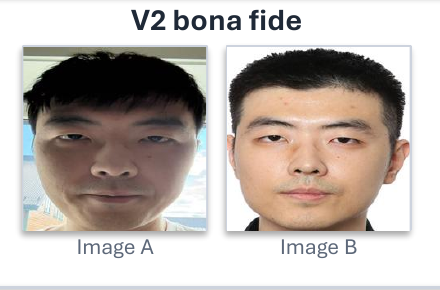}

\raggedright
\textbf{Bona fide, correct.} Fused morph risk score: $10/100$.
\textbf{Audit rationale:} The response reports consistent facial geometry and
landmarks around the eyes, nose, and mouth, uniform skin texture, and no
visible blending, ghosting, local averaging, or boundary artefacts. The
reported reliability is high.
\end{minipage}
\hfill
\begin{minipage}[t]{0.48\textwidth}
\centering
\includegraphics[width=0.62\linewidth]{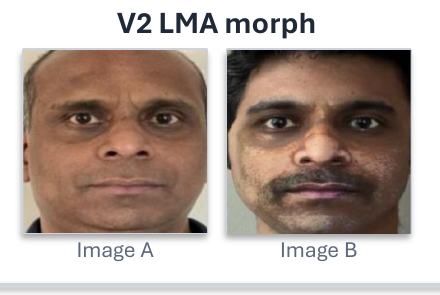}

\raggedright
\textbf{LMA, false negative.} Fused morph risk score: $5/100$.
\textbf{Audit rationale:} The response notes differences in eye shape and
spacing, nose width, and nostril shape, but identifies no visible blending,
ghosting, or boundary artefacts. It therefore predicts bona fide despite the
reported geometric differences.
\end{minipage}

\caption{Representative outputs for the selected recursive prompt
\texttt{G1\_P014}. The LMA pair is missed despite reported geometric
differences because the response gives greater decision weight to the absence
of visible surface artefacts.}
\label{fig:supp_qualitative_v2}
\end{figure*}

The two prompts exhibit the same principal failure mode. Both accept LMA
pairs when surface appearance remains natural and explicit blending artefacts
are weak. For the Version~2 example, the response reports geometric
differences but still rejects the attack hypothesis. This suggests that the
selected instructions give substantial weight to visible artefact absence and
insufficient weight to identity inconsistency. The observation is consistent
with the quantitative LMA results, but it remains an interpretation of the
reported audit rationale rather than a verified account of the model's causal
features.

\FloatBarrier
\section{Response Budget and Reliability}
\label{supp:reliability}

The three saved responses per pair permit an offline response budget ablation.
For a one-response or two-response budget, MACER is averaged over all subsets
of the corresponding size. Each subset uses the OR decision rule. The full
three-response condition uses all saved outputs.

\begin{table*}[t]
\centering
\caption{Response budget and reliability. MACER for one and two responses is
averaged over all subsets of the stated size. FTA and PFR are response level
rates over 480 responses per prompt; inconsistency is measured over pairs with
at least two valid decisions. Lower is better.}
\label{tab:supp_repeat_ablation}
\setlength{\tabcolsep}{6pt}
\renewcommand{\arraystretch}{1.05}
\begin{tabular}{lccc|ccc}
\toprule
& \multicolumn{3}{c|}{\textbf{MACER (\%)}} &
\multicolumn{3}{c}{\textbf{Reliability (\%)}}\\
\textbf{Prompt} & \textbf{1 response} & \textbf{2 responses} &
\textbf{3 responses} & \textbf{FTA} & \textbf{PFR} &
\textbf{Inconsistency}\\
\midrule
Manual & 72.50 & 71.11 & 70.00 & 0.00 & 0.00 & 3.13\\
\texttt{P033} & 23.33 & 20.00 & 19.17 & 0.00 & 0.63 & 6.88\\
\texttt{G1\_P014} & 32.78 & 31.94 & 31.67 & 0.00 & 0.00 & 1.88\\
\bottomrule
\end{tabular}
\end{table*}

Repeated responses provide the largest improvement for \texttt{P033}, reducing
MACER from $23.33\%$ with one response to $19.17\%$ with three responses.
Most of the gain is obtained with two responses. The manual and recursive
prompts improve by $2.50$ and $1.11$ percentage points, respectively. No FTA
occurs. Prompt \texttt{P033} produces three parse failures among 480 calls,
and each affected pair retains at least one valid response. Its higher
inconsistency therefore coexists with a lower fused MACER: OR fusion converts
occasional attack indications into an attack-sensitive final decision.

\section{Subject Level Uncertainty and Paired Tests}
\label{supp:statistics}

Pair observations are clustered because each test subject contributes three
attack pairs. The primary uncertainty analysis therefore resamples the 40 test
subjects with replacement for $100{,}000$ bootstrap iterations and recomputes
MACER from all attack pairs associated with the sampled subjects.

\begin{table}[t]
\centering
\caption{Paired MACER comparison with \texttt{P033}. Positive differences
indicate higher error for the comparison prompt. Intervals resample subjects.}
\label{tab:supp_statistics}
\setlength{\tabcolsep}{4.5pt}
\renewcommand{\arraystretch}{1.05}
\begin{tabular}{lcc}
\toprule
\textbf{Comparison} & \textbf{Difference} & \textbf{95\% interval}\\
\midrule
Manual $-$ \texttt{P033} & 50.83 & [43.33, 58.33]\\
\texttt{G1\_P014} $-$ \texttt{P033} & 12.50 & [6.67, 18.33]\\
\bottomrule
\end{tabular}
\end{table}

Prompt \texttt{P033} has fewer missed attacks than the manual prompt for 37
of 40 subjects and ties it for the remaining three. It has fewer missed
attacks than \texttt{G1\_P014} for 13 subjects and ties it for 27; no subject
favours the comparison prompt in either analysis. Exact McNemar tests on fused
pair decisions provide
supplementary pair level evidence. There are 61 discordant pairs favouring
\texttt{P033} over the manual prompt and none in the reverse direction
($p=8.67\times10^{-19}$). There are 15 discordant pairs favouring
\texttt{P033} over \texttt{G1\_P014} and none in the reverse direction
($p=6.10\times10^{-5}$). The subject bootstrap remains the primary analysis
because it accounts directly for repeated observations from one identity.

\section{Recursive Prompt Optimisation Diagnostics}
\label{supp:optimisation}

\begin{table}[t]
\centering
\caption{Validation security loss across recursive generations. Lower is
better.}
\label{tab:supp_generations}
\setlength{\tabcolsep}{5pt}
\renewcommand{\arraystretch}{1.05}
\begin{tabular}{lcc}
\toprule
\textbf{Generation} & \textbf{Prompts} & \textbf{Best}\\
\midrule
G0 & 30 & 1.10\\
G1 & 15 & 1.10\\
G2 & 15 & 1.20\\
\bottomrule
\end{tabular}
\end{table}

The best validation security losses are $1.10$, $1.10$, and $1.20$ for
Generations~0, 1, and 2, respectively. Generation~1 therefore matches, but does
not improve upon, the best Generation~0 prompt, while Generation~2 produces a
higher minimum loss. The validation split contains 30 attack pairs, so one additional missed attack
changes MACER by $3.33$ percentage points and changes
$\mathcal{L}_{\mathrm{sec}}$ by $0.10$. The difference between the best G0 and
G2 values therefore corresponds to one attack decision. This coarse
resolution limits the ability of recursive search to distinguish closely
performing prompts.

The surviving lineages show regression towards the validation mode: a weaker
G0 ancestor improves from a loss of $1.50$ to $1.10$, while the strongest G0
ancestor degrades from $1.10$ to $1.20$. In a post hoc test comparison, the
retained seed \texttt{G0\_P026} obtains $30.83\%$ MACER, slightly below the
$31.67\%$ of \texttt{G1\_P014}. This diagnostic comparison did not influence
prompt selection; G0 prompts are excluded from the reported Version~2
condition because they have not undergone recursive revision.

For the retained Version~2 validation populations, BSCER, FTA, PFR, and
InconsistencyRate are zero. Consequently, the security loss reduces to a
positive multiple of MACER for these candidates. Replacing the MACER
coefficient by $1$, $2$, $5$, or $10$, using equal weights, or ranking by
MACER alone leaves the ordering unchanged, with Kendall $\tau=1.00$, and
selects the same recursive descendant. This analysis does not establish that
the coefficients are universally irrelevant; it shows only that the observed
Version~2 selection is insensitive to them on the present validation split.

\section{Reproducibility, Data Handling, and Scope}
\label{supp:reproducibility}

All roles use the mutable \texttt{gpt-4o} alias. The Generator and Optimizer
Agents use temperature $0.8$, while the Critic Agent and evaluator use
temperature $0$. Because the alias was not pinned and the query dates are not
recoverable, the underlying weights cannot be reconstructed exactly. All
analyses were computed offline from the saved outputs.

After the anonymous review process, the reproducibility package will contain
the 50 Version~1 prompts and rankings; the Generation~0--2 populations and
lineage fields; the frozen \texttt{P033} and \texttt{G1\_P014} records; the
manual instruction; parsing, fusion, metric, bootstrap, and ablation scripts;
de-identified outputs; and a README specifying the response schema, split
sizes, model alias, temperatures, and procedure.

The release excludes API credentials, raw facial images, image paths, and
direct subject identifiers. External processing of identifiable facial images
must comply with the consent, access, retention, and processing conditions of
the source collection and deployment jurisdiction. The reported system is
intended only to assist trained document inspection personnel and should not
serve as the sole basis for a consequential identity decision.

{
    \scriptsize
    \bibliographystyle{ieeenat_fullname}
    \bibliography{FaceMorph_Refereces}
}